\pdfoutput=1

\documentclass[11pt]{article}

\usepackage[]{acl}

\usepackage{times}
\usepackage{latexsym}

\usepackage[T1]{fontenc}
\DeclareUnicodeCharacter{0254}{\textopeno} 

\usepackage[utf8]{inputenc}

\usepackage{microtype}

\usepackage{inconsolata}
\usepackage{CJKutf8}

\usepackage{amssymb}

\usepackage{subcaption}

\usepackage{array}
\usepackage{graphicx}   
\usepackage{hyperref}   
\usepackage{booktabs}   
\usepackage{multirow}   
\usepackage{tabularx}

\definecolor{lightgreen}{RGB}{144,238,144}
\definecolor{lightred}{RGB}{255,187,187}
\definecolor{lightorange}{RGB}{255,223,155}

\definecolor{iosame}{RGB}{142,209,196}
\definecolor{ioother}{RGB}{191,160,238}
\definecolor{ioeng}{RGB}{144,205,255}

\title{
Evaluating the Elementary Multilingual Capabilities \\ of Large Language Models with \textsc{MultiQ}
}

\author{Paul Röttger$^{1}$, Hannah Rose Kirk$^{2}$, Bertie Vidgen$^{2}$ \\ \textbf{Giuseppe Attanasio}$^{1}$\textbf{,} \textbf{Federico Bianchi}$^{3}$ \and \textbf{Dirk Hovy}$^{1}$ \\ \\ {$^1$MilaNLP, Bocconi University, $^2$University of Oxford, $^3$Stanford University}}

\author{Carolin Holtermann\textsuperscript{1*}, Paul Röttger\textsuperscript{2*}, Timm Dill\textsuperscript{1}, Anne Lauscher\textsuperscript{1} \\
\textsuperscript{1}Data Science Group, University of Hamburg, Germany \\
  \textsuperscript{2}Bocconi University, Italy \\
  \texttt{carolin.holtermann@uni-hamburg.de}}

\begin{document}

\maketitle
\def\thefootnote{*}\footnotetext{Equal contribution}\def\thefootnote{\arabic{footnote}}

\begin{abstract}
Large language models (LLMs) should benefit everyone, including a global majority of non-English speakers.
However, most LLMs today, and open LLMs in particular, are often intended for use in just English (e.g.\ Llama2, Mistral) or a small handful of high-resource languages (e.g.\ Mixtral, Qwen).
Recent research shows that, despite limits in their intended use, people prompt LLMs in many different languages.
Therefore, in this paper, we investigate the basic multilingual capabilities of state-of-the-art open LLMs \textit{beyond their intended use}.
For this purpose, we introduce \textsc{MultiQ}, a new silver standard benchmark for basic open-ended question answering with 27.4k test questions across a typologically diverse set of 137 languages.
With \textsc{MultiQ}, we evaluate language fidelity, i.e.\ whether models respond in the prompted language, and question answering accuracy.
All LLMs we test respond faithfully and/or accurately for at least some languages beyond their intended use.
Most models are more accurate when they respond faithfully.
However, differences across models are large, and there is a long tail of languages where models are neither accurate nor faithful.
We explore differences in tokenization as a potential explanation for our findings, identifying possible correlations that warrant further investigation.

\end{abstract}

\section{Introduction}
Languages other than English remain underrepresented and underserved by state-of-the-art language technologies, posing a barrier to equal and inclusive AI~\citep{bender2011achieving, joshi-etal-2020-state}.
While proprietary large language models~(LLMs) like GPT-4 \citep{openai2023gpt4} may answer questions and follow instructions in many different languages, even the best and most popular open LLMs are much more restricted in their language coverage:
Llama2-chat~\citep{touvron2023llama2}, for example, is \emph{``intended for commercial and research use in English''}.\footnote{\url{https://huggingface.co/meta-llama/Llama-2-7b-hf}}
Yi is \emph{``bilingual''} in English and Chinese\footnote{\url{https://huggingface.co/01-ai/Yi-34B-Chat}}, and Mistral-7b-instruct~\citep{jiang2023mistral} \emph{``only works in English''}.\footnote{\url{https://mistral.ai/news/la-plateforme/}}

\begin{figure}
    \centering
    \includegraphics[width=8cm]{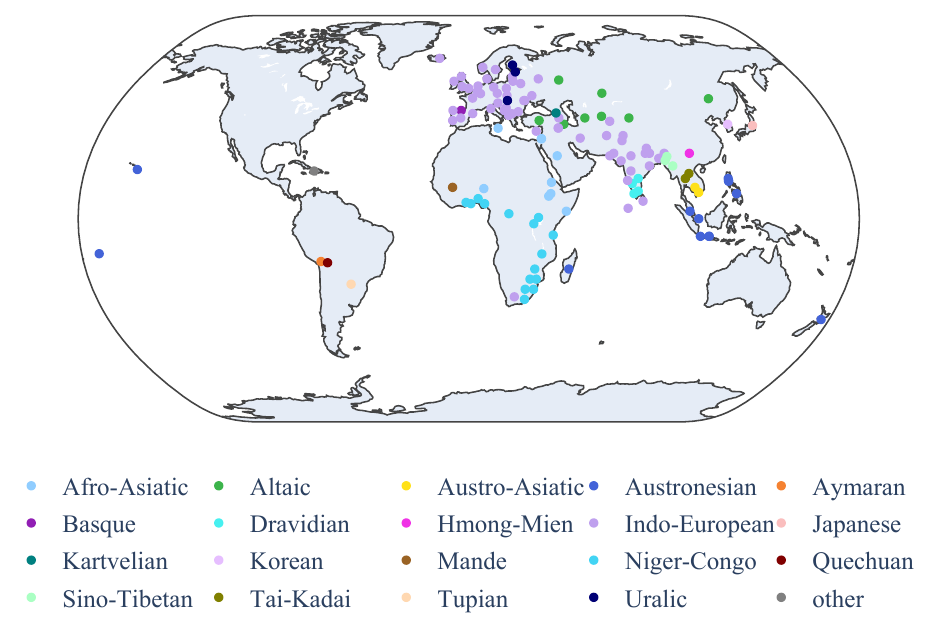}
    \caption{The 137 languages covered in our \textsc{MultiQ} question dataset. We show their geographic location according to the WALS database and indicate their corresponding language family through colors.}
    \label{fig:multiq}
\end{figure}

Even though most open LLMs are restricted in their \emph{intended} use to one or a handful of languages, datasets of real-world LLM usage show that people prompt LLMs in many different languages, often beyond their intended use \citep{ouyang2023shifted,zhao2024inthewildchat, zheng2024lmsys}.
This has motivated initial research into the multilingual capabilities of monolingual models \citep{armengol-estape-etal-2022-multilingual, lai-etal-2023-chatgpt}.
However, this research has mostly focused on older proprietary LLMs and on a relatively small number of languages and/or specific tasks.  

In this paper, we investigate the basic multilingual capabilities of a variety of state-of-the-art chat-optimized open LLMs across a typologically diverse set of 137 languages.
Specifically, we ask two main research questions that correspond to two dimensions of multilingual capability:
{1)~What is the \textbf{multilingual language fidelity} of current chat-optimized open LLMs}, and 
2)~What is the \textbf{multilingual question answering (QA) accuracy} of current chat-optimised open LLMs?
Language fidelity describes the ability to respond to prompts in the prompted language.
QA accuracy describes the ability to give correct answers to open-ended questions, in the prompted language or any other.
An ideal multilingual model would give answers that are both faithful and correct.

To answer our two research questions, we introduce \textsc{MultiQ}, a new silver standard benchmark for basic open-ended question answering comprising 27,400 test prompts across 137 typologically diverse languages.
We create \textsc{MultiQ} by compiling 200 English questions that are simple yet realistic and diverse, and translating them automatically to 136 other languages. 
We evaluate QA accuracy on \textsc{MultiQ} using a GPT-4 classifier and language fidelity using GlotLID \citep{kargaran2023glotlid}.
\textsc{MultiQ} is a silver standard because automated translation and evaluation introduce some noise into the results.
However, we validate through expert annotation that this noise is likely small, thus demonstrating that \textsc{MultiQ} can provide valuable evidence on basic multilingual capabilities.
Concretely, we use \textsc{MultiQ} to make four main findings:

\begin{enumerate}
    \itemsep0em 
    \item \textbf{Language Fidelity}:
    While some open models (e.g.\ Llama2) mostly respond in English regardless of the input language, other models (e.g.\ Mistral) respond faithfully despite their intended use being monolingual.
    \item \textbf{QA Accuracy}:
    On \textsc{MultiQ}, all models tend to perform best in English, with some performing similarly well in up to 20 other languages (e.g.\ Mixtral).
    Across models, there is a long tail of languages with very poor accuracy.
    \item \textbf{Positive Interaction}: Increased language fidelity appears to positively impact answer accuracy, since model answers that match the prompt language tend to be more accurate.
    \item \textbf{Tokenization as (Partial) Explanation}: Models tend to achieve higher accuracy on languages they can tokenize into subwords instead of characters or ASCII tokens.
\end{enumerate}

We publish all data and code at \url{https://github.com/paul-rottger/multiq}

\section{The \textsc{MultiQ} Dataset} \label{sec:multiq}




\textsc{MultiQ} is a collection of 27,400 simple open-ended questions across 137 typologically diverse languages.
The questions cover different topics ranging from algebra to geography to astronomy.
Questions in each language are parallel to each other.
The open-ended question format is consistent with real-world LLM usage.
Additionally, the open-endedness minimizes the likelihood of correct answers given by chance.

\subsection{Dataset Creation}
We created \textsc{MultiQ} in two steps.
First, we compiled an initial set of English questions. 
Second, we automatically translated these prompts into 136 typologically diverse languages.

For the \textbf{initial English questions}, we used two different sources to increase question diversity.
\textbf{a)}~We collected 100 questions from the LMSYS-Chat-1M dataset \citep{zheng2024lmsys}, which catalogs real-world user interactions with LLMs.
Specifically, we sampled all single-sentence English-language and sorted them by frequency.
Then, we manually selected from the top until we reached 100 questions.
\emph{This portion of our data directly reflects real-world LLM usage.}
\textbf{b)} We manually created another set of 100 questions
evenly spread across 10 different subjects at elementary to middle school level (e.g.\ mathematics and geography).
To maximize the diversity of the questions, we prompted GPT-4 to provide us with a set of simple and clear questions with simple and clear answers for each of the subjects.
We then iterated and manually selected questions until we reached 10 questions per subject.
\emph{This portion of our data expands \textsc{MultiQ}'s topical coverage.} 

In both the LMSYS and the GPT-4 portions of our data, we manually selected only questions that are simple, factual, and target common knowledge.
This is because with \textsc{MultiQ} we want to test \emph{basic} multilingual capabilities, not complex reasoning.
Questions must also have unambiguous answers that are culturally and temporally invariant.
This is to minimize discrepancies introduced by translation as well as temporal degradation of our dataset. Table \ref{tab:prompt_examples} shows English example prompts.

\begin{table}[t]
\small
    \centering
    \begin{tabular}{l|l}
    \toprule
    \textbf{Source}  & \textbf{Example Prompts} \\
    \midrule
     \multirow{3}{*}{LMSYS}  & \emph{Was the year 2000 a leap year?} \\ 
     & \emph{What is 2 + 2 * 3?} \\
     & \emph{How many feet does a chicken have?} \\
    \midrule
    \multirow{3}{*}{GPT-4}  &  \emph{What is the chemical formula for water?} \\
    & \emph{Who was the first Emperor of China?} \\
     & \emph{What is a galaxy?} \\
    \bottomrule
    \end{tabular}
    \caption{Examples of English questions covered in \textsc{MultiQ}. We present three prompts from each source (LMSYS and GPT-4) covering different domains.}
    \label{tab:prompt_examples}
\end{table}

To \textbf{translate our English questions} into other languages, we used the v3 Google Translate API.%
\footnote{\url{https://translation.googleapis.com/v3/}}
Specifically, we translated the 200 initial English questions into all 136 other languages covered by the API as of February 2024, resulting in a total of 27,400 parallel questions in \textsc{MultiQ}.
The decision to use automated translation was driven by the constraints of our research budget, which made manual translation infeasible.
The benefit of automated translation is that we can cover many more languages.
Next, we discuss how we validated the quality of the translations, and demonstrate the typological diversity of the 137 languages we cover.

\subsection{Validation of Translations}
We asked native speakers to annotate the correctness of the 200 translated \textsc{MultiQ} questions for 19 languages: Arabic, Catalan, Chinese, Farsi, French, German, Hindi, Indonesian, Italian, Japanese, Korean, Spanish, Tagalog, Russian, Spanish, Quechua, Ukrain, Urdu and Xhosa. Across these languages, annotators marked an average of 91.6\% of translations as correct.
Translations were least accurate for Tagalog at 60.0\% and Arabic at  82.2\%, while the translations for Italian and German were the most accurate, at 99.0\%.
We present the full results in the Appendix \ref{app:multiq}.
Qualitatively, several annotators stated that some translations, while accurate in content, tended to be literal, word-for-word translations rather than natural expressions.
Overall, the automated translation introduces some noise into \textsc{MultiQ}, but our validation results suggest that the amount of noise is limited.
This is why we frame \textsc{MultiQ} as a \emph{silver standard} benchmark that can provide meaningful insights into basic multilingual capabilities, even if exact results on individual test cases may not be perfectly reliable.

\subsection{Typological Diversity}
\textsc{MultiQ} covers a total of 137 languages.
To demonstrate their typological diversity, we follow best practices suggested by \citet{ploeger2024typological}, analyzing both between-language distances as well as overall typological feature coverage.

\begin{figure}
    \centering
    \includegraphics[width=\linewidth]{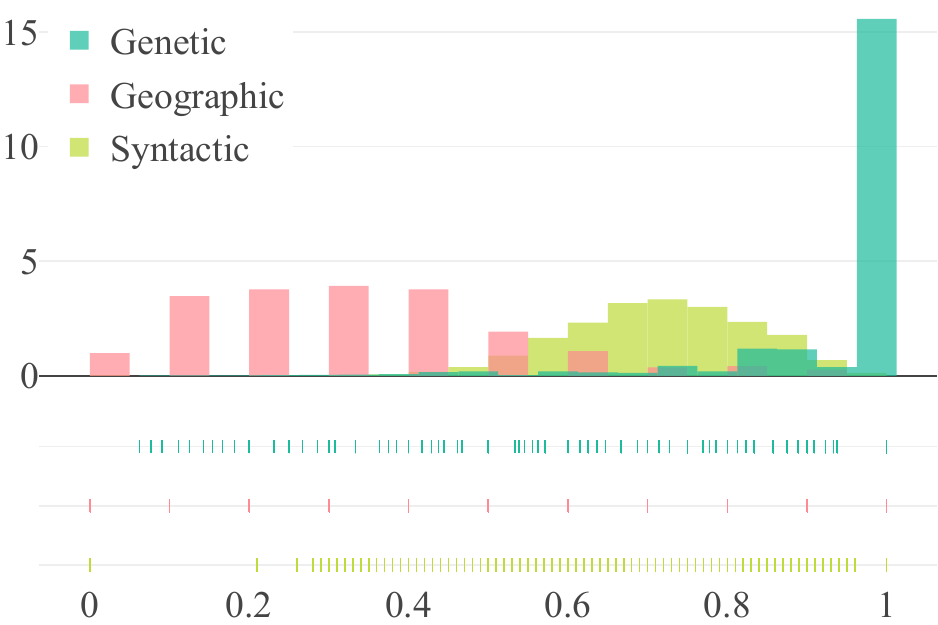}
    \caption{Distributions of the pairwise \texttt{lang2vec} distances for each language pair present in \textsc{MultiQ}.}
    \label{fig:typology}
\end{figure}

To estimate \textbf{language distances}, we use the \texttt{lang2vec} toolkit \citep{littell2017uriel} which contains precomputed language distances based on the typological language information queried from the URIEL knowledge base.\footnote{\url{https://www.cs.cmu.edu/~dmortens/projects/7_project/}}
Following \citet{ploeger2024typological}, we calculate the distances between all pairs of languages in our dataset that have at least 5\% coverage in the URIEL vectors. Figure \ref{fig:typology} shows the distribution of pairwise geographic, syntactic, and genetic distances of all covered languages. We find that the languages in \textsc{MultiQ} cover a wide range of typologically similar and distant language pairs with an expected high skewness of genetic distance complementing previous research \citep{ploeger2024typological}.  We can therefore confidently speak of a high typological diversity of our dataset.

Next, we calculate the \textbf{typological feature coverage} of the 137 languages in \textsc{MultiQ}, using language features provided in the Grambank database \citep{grambank}.
For this purpose, we map the language IDs in \textsc{MultiQ} (obtained from the Google Translate API) to the Glottoids used by Grambank.%
\footnote{For 14 languages in our dataset there is no matching entry in Grambank. To avoid incorrect mapping, we do not assign them manually and exclude these languages from the calculation}
Taken together, the languages in \textsc{MultiQ} cover at least 95.4\% of the typological features recoded in Grambank.
This underlines the typological diversity of our dataset.

Finally, we classify each language by its \textbf{language family} using the World Atlas of Language Structure (WALS).\footnote{\url{https://wals.info/languoid}}
In total, the 137 languages in  \textsc{MultiQ} belong to 20 different language families.%
\footnote{For languages that cannot be found in WALS (e.g.\ Corsican), we manually look up the language family in Grambank.}
For additional details on the languages in our dataset, see Appendix \ref{app:multiq}.

\section{Experiments and Results}\label{sec:experiments}
Using \textsc{MultiQ}, we can now answer our two main research questions regarding the \textbf{multilingual language fidelity} and \textbf{multilingual QA accuracy} of current chat-optimized open LLMs. We first separately assess fidelity and accuracy, and then evaluate their relationship.

\subsection{Overall Experimental Setup}

\paragraph{Models}
We test six open-access LLMs that are both popular and competitive in performance with other state-of-the-art models as measured on standard (English-language) benchmarks such as the LMSys Leaderboard\footnote{\url{https://huggingface.co/spaces/lmsys/chatbot-arena-leaderboard}} and AlpacaEval.\footnote{\url{https://tatsu-lab.github.io/alpaca_eval/}} Specifically, the 7B, 13B and 70B versions of Llama2-Chat \citep{touvron2023llama2}, the 7B Mistral-Instruct-v0.1 \citep{jiang2023mistral}, the 8x7B Mixtral-Instruct-v0.1 \citep{jiang2024mixtral} and the 7B Qwen1.5-Chat model \citep{bai2023qwen}.
We test three sizes of Llama2 to evaluate scaling on \textsc{MultiQ}.
Llama2 and Mistral are explicitly intended for English use only, whereas Mixtral and Qwen are explicitly multilingual:
Mixtral \emph{``handles English, French, Italian, German and Spanish''}, while Qwen offers unspecified \textit{``multilingual support''}.

\paragraph{Inference}
We run all models on two A100 GPUs using the \texttt{simplegen} Python library \citep{milanlp-2023-simple-generation}.
We use default generation parameters from the \texttt{transformers} library, except for temperature, which we set to 0 to make completions deterministic.
The maximum length of generations is 256 tokens.
We do not use any system prompts.
When prompting models with \textsc{MultiQ} questions, we do not provide any additional context or examples.

\subsection{Language Fidelity} \label{subsec:fidelity}

\begin{figure}
    \centering
    \includegraphics[width=0.48\textwidth]{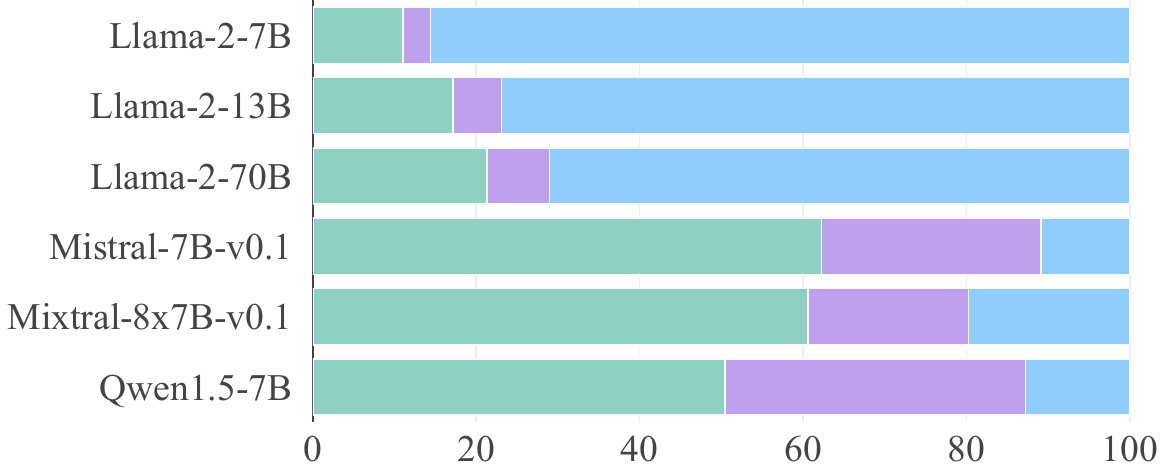}
    \caption{Overall language fidelity. Proportion of model responses (\%) in the \colorbox{iosame}{same language} as the input prompt, in \colorbox{ioeng}{English}, or in \colorbox{ioother}{another language}\textcolor{ioother}. We evaluate the responses of six models for 200 prompts in 135 languages (excl. Dogri \& Meiteilon).}
    \label{fig:io-language}
\end{figure}

We gather responses from all six models described above on the 27,400 questions in \textsc{MultiQ} and then use GlotLID \citep{kargaran2023glotlid} to identify the response language.
GlotLID is an open-source language identification model that supports more than 1,600 languages.
GlotLID returns \texttt{iso\_636\_9} language codes, which we manually map to the language codes in \textsc{MultiQ}.%
\footnote{For 13 languages, several language codes in \textsc{MultiQ} map to just one \texttt{iso\_636\_9} code. For details see Appendix \ref{app:exp_setup}.} 
Two languages in \textsc{MultiQ}, namely Meiteilon (Manipuri) and Dogri, are not supported by GlotLID, so we exclude them from our language fidelity analysis.
Figure~\ref{fig:io-language} shows high-level results on language fidelity, split by how often models responded in the language of the input prompt, in English, or another language.

We find that the Llama2 models show a very low language fidelity, responding predominantly in English, matching its intended use for English only. 
Fidelity increases with scale, but even Llama2 70b, which gives 21.4\% answers in the prompt language, is much less faithful than the other models. Surprisingly, Mistral, also intended for English use only, shows the greatest language fidelity, giving 62.3\% of answers in the prompt language. Mistral is closely followed by Mixtral (60.6\%) and Qwen (50.4\%), which are advertised as having multilingual capabilities. Interestingly, compared to Llama2, the other models more frequently opt neither for English nor the prompt language, but some other language in their response. This effect appears most evident for Qwen.

To confirm the robustness of our findings, we investigate the impact of brief and numerical responses from the models on our results. To this end, we excluded all questions from the \textsc{MultiQ} dataset that required a numerical answer, specifically removing the 10 curated questions of the domain ``math'' as well as 16 questions that were drawn from the LMSYS dataset. We then analyzed the character length of the model responses, noting an average length of 270-670 characters across different models. We also exclude responses shorter than 10 characters from the language fidelity calculation. The overall language fidelity of the models showed only marginal changes in response to these changes, underscoring the minimal influence of brief or numerical responses on our analysis.

\begin{figure}
     \centering
     \begin{subfigure}[b]{0.75\linewidth}
         \centering
         \includegraphics[width=\textwidth]{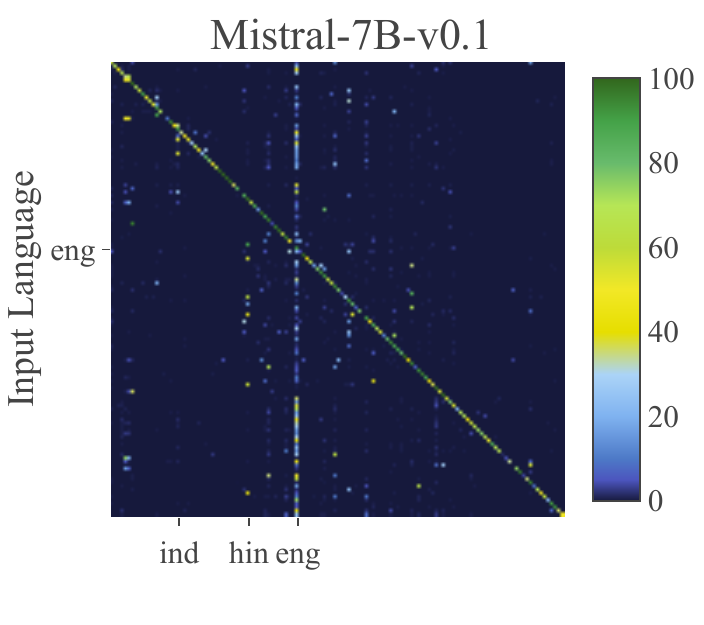}
     \end{subfigure}
     \hfill
     \begin{subfigure}[b]{0.75\linewidth}
         \centering
         \includegraphics[width=\textwidth]{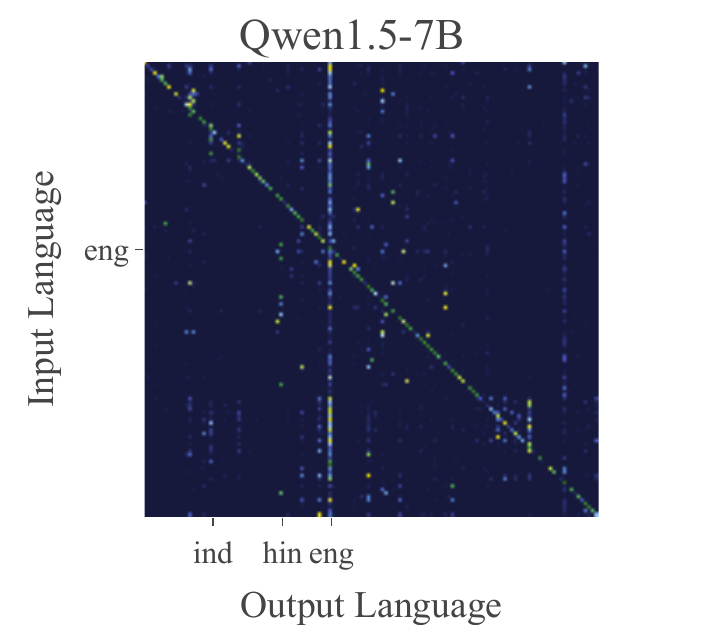}
     \end{subfigure}
        \caption{Granular language fidelity. Correlation matrices illustrating the relationship between input prompt and model response languages, shown as percentages. Axis ticks are selectively labeled for better visualization.}
        \label{fig:io_matrix_plots}
\end{figure}

Next, we conduct a more fine-grained analysis of model answers in ``another language'', i.e.\ neither the input language nor English. For this, we focus on the three models responding in another language more than 10\% of the time (Qwen, Mistral, Mixtral). Of these, Mistral demonstrates the highest level of diversity among the languages it responds in. GlotLID identifies more than 360 distinct languages in Mistral's responses, compared to around 240 languages for Mixtral and Qwen. We also examine the correspondence between the language family of input and output language. While Mixtral provides a response in the same language family at least 68.7\% of the time, if not in an exact match or responding in English, Qwen demonstrates this behavior only 60.6\% of the time, while Mistral does so for 54.6\% of prompts. Note, however, that for 6\% of the answers of Mistral and 7\% of the answers of Mixtral, the language family of the response languages could not be determined and is thus classified as 'unknown'.
Additionally, the distribution of language families in \textsc{MultiQ} is highly skewed, with 45.3\% of the languages belonging to the 'Indo-European' family, which spans a very broad range of languages from Irish to Turkish.%

Finally, we analyze the relationship between the language of the prompt and the frequency of the respective response language of the models.
Figure \ref{fig:io_matrix_plots} shows correlation matrices for Mistral and Qwen.
Corresponding matrices for the other models are shown in Appendix \ref{app:language_fidelity}. We find that Hindi is the most frequently selected language outside of the input language or English, closely followed by Indonesian. For Qwen, for example, these languages make up 20.4\% of the ``another language'' category. This is visible by the thin yellow vertical lines leading to the axis ticks \textit{ind} and \textit{hin} in the Figure for both models respectively. A potential explanation may be that the models lack support for numerous languages from India and Indonesia, thus treating Hindi and Indonesian as some kind of 'fallback' languages for the wider language area. For example, the models often respond to languages such as Malay and Javanese in Indonesian, and languages such as Maithili, Konkani, and Bhojpuri in Hindi. We also observe this phenomenon for smaller European languages. For example, the three models respond to questions in Croatian mostly in Italian, to those in Luxembourgish mostly in German, and those in Galician mostly in Portuguese. These observations underline the importance of improving multilingual models in the language coverage of low-resource languages.


\begin{table}[t]
\small
    \centering
    \begin{tabular}{ll|ccc}
    \toprule
        \textbf{Model} & \textbf{Label} & \textbf{P} & \textbf{R} & \textbf{F1}\\
        \midrule
        \multirow{2}{*}{Mistral-7B} & Incorrect A. & 0.97 & 1.00 & 0.98\\
        & Correct A. & 0.98 & 0.87 & 0.92\\
        \midrule
        \multirow{2}{*}{Qwen1.5-7B} & Incorrect A. & 0.88 & 0.99 & 0.93\\
        & Correct A. & 0.94 & 0.61 & 0.74\\
        \midrule
        \multirow{2}{*}{Llama-2-7b} & Incorrect A. &  0.92 & 1.00 & 0.96\\
        & Correct A. & 1.00 & 0.74 & 0.85\\
        \midrule
        \multirow{2}{*}{Mixtral-8x7B} & Incorrect A. &  0.82 & 0.98 & 0.89\\
        & Correct A.&  0.97 & 0.71 & 0.82\\
        \bottomrule
    \end{tabular}
    \caption{Evaluation of the GPT-4 classifier that we use to assess answer accuracy on \textsc{MultiQ}. We show Precision (P), Recall (R), and F1-Score of GPT-4 judgments on responses from four models to the same sample of 282 questions covering all languages, which were annotated by humans for whether they are correct or not.}
    \label{tab:validation_metrics}
\end{table}

\subsection{Question Answering Accuracy}\label{subsec:answeracc}

Next, we assess how often models give correct answers, regardless of whether the language of this answer matches the input prompt or not.

\paragraph{Automated Evaluation}
Since questions in \textsc{MultiQ} are open-ended and answers can come in many languages, we need a flexible method for evaluating accuracy.
We find that a carefully crafted prompt to GPT-4, which checks a given model answer against the English version of the question from \textsc{MultiQ}, serves this purpose well.%
\footnote{See Appendix \ref{app:exp_setup} for the prompt template.}

To evaluate the reliability of our automated evaluation method, we tasked two independent human annotators to label model responses from four of our models for the same 282 randomly selected prompts covering at least two questions per language, as correct or incorrect.%
\footnote{We exclude the 13B and 70B versions of Llama2 from the QA accuracy analysis for reasons of clarity, and because they resemble the results of Llama2 7B.}
Disagreements between the annotators, which occurred for no more than four responses per model, were resolved in discussions with one of the authors.
We find that the accuracy of our automated evaluation, as measured against the human labels, is very high across models (see Table~\ref{tab:validation_metrics}). For all models, the automated evaluation tends to be very precise on correct answers, but less so on incorrect answers. This means that the automated evaluation will likely underestimate the proportion of correct model answers across languages. Overall, automated evaluation, like automated translation, introduces noise to our silver standard \textsc{MultiQ} benchmark, but we find the amount of noise to likely be small.

\paragraph{Results}

\begin{figure*}
    \centering
    \includegraphics[width=\textwidth]{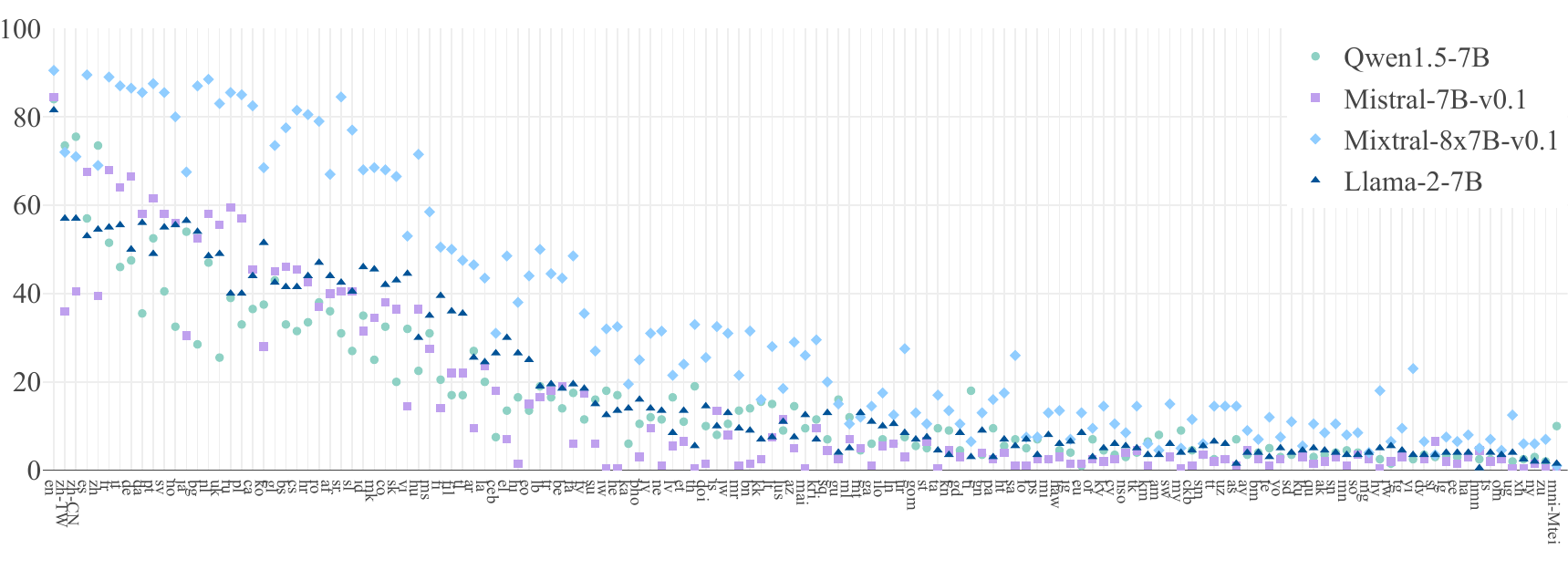}
    \caption{Answer accuracy on \textsc{MultiQ} in proportion (\%) of correctly answered questions per language. We compare four models of the same size across 137 languages and sort the results by median accuracy.}
    \label{fig:answer_accuracy}
\end{figure*}

\begin{table}[]
\small
    \centering
    \begin{tabular}{l|c|c|c|c|c}
    \toprule
    Model & All & \textsc{en} & $\blacktriangle$10 & $\blacktriangle$20 & $\blacktriangle$50 \\
    \midrule
    Qwen (7B) & 16.6 & 84.0 & 61.6 & 50.2 & 34.3 \\
    Mistral (7B) & 15.4 & 84.5 & 64.6 & 56.6 & 37.6 \\
    Mixtral (8x7B) & \textbf{33.8} & \textbf{90.5} & \textbf{87.7} & \textbf{84.8} & \textbf{67.4} \\
    Llama2 (7B)& 19.2 & 81.5 & 58.4 & 53.9 & 41.4\\
    Llama2 (13B)& 23.4 & 82.0 & 66.4 & 62.7 & 49.6\\
    Llama2 (70B)& 29.1 & 90.5 & 80.7& 76.5 & 61.2\\
    \bottomrule
    \end{tabular}
    \caption{QA accuracy on \textsc{MultiQ} (\%).
    We show accuracy overall, on English questions, and on the top($\blacktriangle$) 10, 20 and 50 best-performing languages for each model.
    Highest accuracy across models is \textbf{bold}.}
    \label{tab:answer_acc}
\end{table}

Based on our automated evaluation, we calculate the proportion of correctly answered questions for each model across all 137 languages.
Table~\ref{tab:answer_acc} shows overall results, and Figure~\ref{fig:answer_accuracy} shows a breakdown across all languages.
We find that Mixtral is most accurate overall, but also across most individual languages covered in \textsc{MultiQ}.
Mistral shows the lowest overall accuracy across all languages, with a long tail distribution starting to decrease after the best-performing 20 languages. Moreover, a direct comparison between the results of Mixtral and Qwen shows that they achieve very similar results on Chinese, although only Qwen was explicitly trained in Chinese \citep{bai2023qwen}.

\begin{table}[]
\small
    \centering
    \begin{tabular}{l|c|c|c}
    \toprule
    Model & \colorbox{iosame}{Same} & \colorbox{ioeng}{English} & \colorbox{ioother}{Other} \\
    \midrule
    Qwen (7B) & \textbf{21.5} & 11.4 & 11.7 \\
    Mistral (7B) & \textbf{17.1} & 14.6& 11.8 \\
    Mixtral (8x7B) & 35.0 & \textbf{37.0} & 26.7 \\
    Llama2 (7B) & 44.7 & 14.9 & \textbf{46.5} \\
    Llama2 (13B) & \textbf{51.6} & 15.1 & 49.6 \\
    Llama2 (70B) & \textbf{61.4} & 17.3 & 48.2 \\
    \bottomrule
    \end{tabular}
    \caption{QA accuracy on \textsc{MultiQ} (\%) split by language fidelity, i.e.\ which language models answered in (see Figure~\ref{fig:io-language}).
    Highest accuracy per model is \textbf{bold}.}
    \label{tab:fid_acc}
\end{table}

\subsection{Language Fidelity vs Answer Accuracy}
Finally, we combine the results of \S\ref{subsec:fidelity} and \S\ref{subsec:answeracc} to assess the relationship between language fidelity and QA accuracy. Table \ref{tab:fid_acc} shows the mean accuracy grouped by response language category. 
We find that Llama2, despite its overall low language fidelity, shows strong QA accuracy.  Especially when answering in the same language as prompted, it gives a correct answer in almost half of the cases (44.7\%). Scaling the model further improves accuracy, with the 13B and 70B variants achieving higher correct answer rates of 51.6\% and 61.4\%, respectively, when answering in the same language as prompted. Qwen and Mistral show a similar pattern, i.e.\ higher answering accuracy when answering in the prompt language with 21.5\% and 17.1\%. However, their overall accuracy is quite low, especially when compared to their initial high language fidelity. Strikingly, we find significant differences across models that ought to be similar due to their intended use for English, i.e.\ Llama2 demonstrating low fidelity yet high accuracy, in contrast to Mistral, which exhibits the reverse pattern. A qualitative analysis of Mistral's answers reveals that the model often merely repeats the questions it was asked, which is technically faithful but never an accurate answer (see Table \ref{tab:fidelity_issues}). Only for Mixtral is this pattern not apparent, with the highest accuracy of 37\% being achieved for responses in English. Overall, we find that if models answer in the same language as the prompted language, they tend to be more accurate than if they respond in English. The only exception of this pattern is Mixtral. Therefore, our results suggest that increased language fidelity may positively impact QA accuracy.






\section{Tokenization and Multilinguality} \label{subsec:tokenization}
Our results show significant variations in terms of language fidelity, QA accuracy, and the relationship between them, across the models we test with \textsc{MultiQ}.
This prompts us to investigate what factors may explain these differences. 
We focus on the differences between Mistral and Llama2 7B, which are particularly surprising given that both models are intended for use in English only.
Prior research highlights the significant roles of tokenization and training data in multilingual capability \citep{dufter-schutze-2020-identifying, clark-etal-2022-canine, petrov2023language}.
Since there is little to no public information on the training data of the models that we test, we focus on their tokenizers.

\paragraph{Background: Byte Pair Encoding} Both Llama2 and Mistral employ a tokenizer that uses byte pair encoding \citep[BPE,][]{sennrich-etal-2016-neural}, streamlining sequence encoding by minimizing token count through identifying common subwords. This allows for frequent sequences to be represented through subword tokens, whereas rare or unseen sequences make the model default to individual characters or, failing that, ASCII code tokens.

\paragraph{Unique Tokens in MultiQ} First, we evaluate how different tokenization strategies impact prompt tokenization in \textsc{MultiQ}. We observe a significant deviation in the number of unique tokens used to represent all prompts in \textsc{MultiQ}: Mistral uses 9,933 unique tokens, contrasting with Llama2's 10,676. This suggests that Llama2 may be less efficient at segmenting the typologically diverse \textsc{MultiQ} dataset into a smaller number of subwords. 

\begin{figure}
    \centering
    \includegraphics[width=\linewidth]{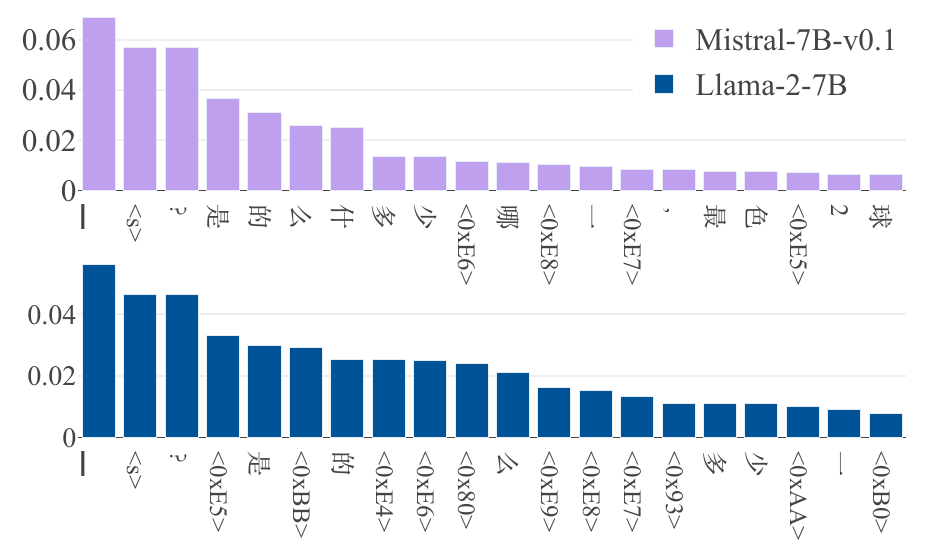}
    \caption{Tokenization analysis. Distribution of the most used unique tokens across all n=200 \textsc{MultiQ} prompts in Chinese (\textsc{zh}), when using the tokenizers of Llama2 (7B) and Mistral (7B).}
    \label{fig:diff_tokenizer}
\end{figure}

\paragraph{Tokenization Strategies} For each model, we group languages into three categories depending on the model's tokenization strategy for the respective language. We develop a heuristic that allows us to classify languages into three tokenization categories: ``ASCII'', ``character'' or ``subword''. We do so by compiling the model's 20 most commonly used tokens for representing each language in \textsc{MultiQ}, removing noise tokens (e.g.\ sentence start) from the list, and then quantifying the prevalence of ASCII tokens and characters based on the token-id ranges they usually occupy.  Languages with over 70\% of tokens in the ASCII or character categories were classified accordingly. For Llama2, out of 137 languages, we find 89 subword, 36 character and 12 ASCII languages. Mistral has 37 character and 9 ASCII languages. The models differ in the tokenization of some symbolic languages such as Chinese (\textsc{zh}, see Figure \ref{fig:diff_tokenizer}): while Mistral mostly uses individual character tokens, Llama2 more frequently resorts to ASCII tokens due to its limited Chinese token vocabulary, which limits its ability to effectively tokenize Chinese language prompts.

\paragraph{Tokenization vs.\ QA Accuracy}
We evaluate the models' average QA accuracy across tokenization categories, finding a clear hierarchy (see Table~\ref{tab:tok_accuracy}): Subword encoding outperforms character and ASCII encodings, with Mistral and Llama2 respectively achieving 20.4\% and 24.0\% accuracy on subword-encoded languages compared to just 6.4\% and 11.4\% on character-encoded languages. Even though our heuristic may introduce some noise in classifying tokenization strategies, we believe that exploring tokenization optimization is a promising direction for multilingual research, which also aligns with the broader discourse on the impact of tokenization on model capabilities. We hope our insights can help motivate further research into alternative language representation strategies \citep[e.g.\ pixel-based models,][]{salesky-etal-2023-multilingual}.

\begin{table}[]
    \centering
    \small
    \begin{tabular}{l|c|c|c}
    \toprule
        Model & Subword & Character & ASCII \\
        \midrule
        Llama2 (7B) & \textbf{24.0} & 11.2 & 8.0 \\
        Mistral (7B) & \textbf{20.4} & 6.4 & 2.1 \\
        \bottomrule
    \end{tabular}
    \caption{QA accuracy (\%) on \textsc{MultiQ} split by tokenization strategy. Highest accuracy per model is \textbf{bold}.}
    \label{tab:tok_accuracy}
\end{table}

\section{Related Work} 
We discuss the related literature with respect to (i)~large multilingual benchmarks, (ii)~explicitly multilingual models, and (iii)~multilingual studies of monolingual models.

\paragraph{Multilingual Benchmarks} Existing multilingual benchmarks primarily target the performance of fully or partially supervised models on collections of standard NLP tasks, like XNLI~\citep{conneau-etal-2018-xnli}. Popular examples are the XTREME benchmarks~\cite[e.g.][]{ruder-etal-2021-xtreme, ruder-etal-2023-xtreme}, and XGLUE~\citep{liang-etal-2020-xglue}. Similarly, researchers also presented benchmarks designed for languages spoken in particular regions, e.g.\ IndicXTREME~\citep{doddapaneni-etal-2023-towards} for Indic languages, Masakhan-NER~\citep{adelani-etal-2022-masakhaner},  and 
TaTA \citep{gehrmann-etal-2023-tata} covering African languages, as well as NusaX~\citep{winata-etal-2023-nusax} for Indonesian languages.
Recently, \citet{ahuja-etal-2023-mega} proposed MEGA for evaluating multilingual generative models, which they use to evaluate LLMs like GPT-4 on a set of standard tasks. In a similar vein, \citet{asai2023buffet} presented BUFFET and benchmark LLMs for few-shot transfer. Another line of work focuses on multilingual QA datasets for reading comprehension, such XQuaD \citet{artetxe-etal-2020-cross}, TyDiQA \citep{clark-etal-2020-tydi} and the Belebele benchmark \citep{bandarkar2023belebele} with up to 122 diverse languages. Concurrent work proposes a new multilingual instruction tuning dataset called Aya \citep{singh2024aya}, which also covers a range of open-ended questions in 114 languages. By comparison, \textsc{MultiQ} contains parallel questions in a larger set of 137 languages covering 95.4\% of Grambank features, which demonstrates its typological diversity and allows the analysis of multilingual LLM behavior at the margins of language coverage. Furthermore, \textsc{MultiQ}'s carefully selected short and simple questions target basic LLM knowledge to test only their multilingual capabilities and not ancillary factors such as complex reasoning.



\paragraph{Multilinguality in Monolingual LLMs} Given the limited availability of open multilingual chat models, we are especially interested in assessing the multilinguality of models intended for English use only. \citet{blevins-zettlemoyer-2022-language} explained this behavior through data contamination: while the vast majority of the pre-training data of those models is English (e.g.\ \textasciitilde93\% for GPT-3~\citeauthor{brown2020language}, according to \citeyear{brown2020language}, and \textasciitilde90\% for Llama-2 according to \citeauthor{touvron2023llama2}, \citeyear{touvron2023llama2}) there are also small portions of non-English content in the pre-training data. In such cases, it appears that the dominant language can help ``unlock'' the models' capabilities for the underrepresented languages~\citep{gogoulou-etal-2022-cross}. Consequently, recent research assesses the multilinguality of several English-centric models, like GPT-3 and ChatGPT~\citep[e.g.][]{zhang-etal-2023-dont, lai-etal-2023-chatgpt, armengol-estape-etal-2022-multilingual, winata-etal-2021-language}. However, these prior works focus on standard NLP tasks such as text classification, just one or few models, and on just a few major languages. By contrast, we test the multilingual behavior of six LLMs in a more natural open-ended QA setting. 


\section{Conclusion}
We introduced \textsc{MultiQ}, a new silver standard benchmark for open-ended question answering that covers 137 typologically diverse languages. With \textsc{MultiQ}, we evaluated the basic multilingual capabilities of six current, chat-optimized open LLMs, which are restricted in their intended use to just one or a small handful of languages. Our analysis focused on two key dimensions of multilingual capability -- language fidelity and QA accuracy -- and how they relate to each other.
We found that all LLMs we test respond faithfully and/or accurately for at least some languages beyond their intended use.
Most models are more accurate when they respond faithfully.
However, we found that differences across models are large, and that there is a long tail of languages where models are neither accurate nor faithful.
Finally, we indentified differences in tokenization as a potential explanation for our results.
Overall, we hope that our findings can motivate further research into improving the multilingual capabilities of open LLMs, especially for diverse and under-represented languages, so that language technologies can benefit everyone, regardless of which language they speak.

\section*{Limitations}
Our work comes with several limitations, which have already been partially discussed throughout the paper.
First, we automatically translate our dataset, which introduces noise in the dataset. However, based on manual validation by native speakers, translation quality is high.
Second, the automated evaluation of the models' answering accuracy using GPT-4, as well as response language classification using GlotLID, introduce some noise into the results. Here, too, human validation assures us that these factors minimally impact the results, potentially leading to an underestimation of answer accuracy.
Third, mapping Google Translate IDs to the ISO language codes of GlotLID was not always directly possible, but we excluded possible inaccuracies from our calculations.
Lastly, our analysis intentionally concentrated on basic multilingual capabilities, excluding the assessment of advanced reasoning or formulation skills.

\section*{Ethical Considerations}

\paragraph{Intended Use}
As we emphasized throughout our paper, \textsc{MultiQ} is intended to test \textit{basic} multilingual capabilities.
Therefore, good performance on \textsc{MultiQ} alone should not be used as evidence for an LLM being suitable for specific languages.

\section*{Acknowledgements}
The work of Carolin Holtermann and Anne Lauscher is funded under the Excellence Strategy of the German Federal Government and the States. Paul Röttger is a member of the Data and Marketing Insights research unit of the Bocconi Institute for Data Science and Analysis, and is supported by a MUR FARE 2020 initiative under grant agreement Prot.\ R20YSMBZ8S (INDOMITA). We thank our native-language annotators for their valuable work.

\bibliography{custom}

\clearpage
\section*{Appendix}
\appendix
\section{MultiQ} \label{app:multiq}
In total \textsc{MultiQ} covers 10 different question domains and 20 distinct language families. 

\paragraph{Domains}
\begin{itemize}
    \item chemistry
    \item physics
    \item astronomy
    \item history
    \item maths
    \item geography
    \item art
    \item sports
    \item music
    \item animals
\end{itemize}

\paragraph{Language Families}
\begin{itemize}
    \item Afro-Asiatic (AA)
    \item Altaic (Al)
    \item Austro-Asiatic (AuA)
    \item Austronesian (Au)
    \item Aymaran (Ay)
    \item Basque (B)
    \item Dravidian (D)
    \item Hmong-Mien (HM)
    \item Indo-European (IE)
    \item Japanese (J)
    \item Kartvelian (K)
    \item Korean (Ko)
    \item Mande (M)
    \item Niger-Congo (NC)
    \item Other (O)
    \item Quechuan (Qu)
    \item Sino-Tibetan (ST)
    \item Tai-Kadai (TK)
    \item Tupian (T)
    \item Uralic (U)
     
\end{itemize}

\begin{table}[h]
    \centering
    \begin{tabular}{l|c}
    \toprule
       Language  & Prop. of correct questions\\
       \midrule
       Arabic & 0.822 \\
       Catalan & 0.899 \\
       Chinese & 0.955 \\
       Farsi & 0.890 \\
       French & 0.915\\
       German & 0.995 \\
       Hindi & 0.975 \\
       Indonesian & 0.889 \\
       Italian & 0.990 \\
       Japanese & 0.955 \\
       Korean & 0.97 \\
       Spanish & 0.950 \\
       Tagalog & 0.6 \\
       Quechua & 0.895\\
       Romanian & 0.875 \\
       Russian & 0.97 \\
       Ukrain & 0.935 \\
       Urdu & 0.935\\
       Xhosa & 0.990\\
         \bottomrule
    \end{tabular}
    \caption{Validation Results on \textsc{MultiQ}. We present the proportion of correctly translated prompts for each language assessed by native speakers in the respective language.}
    \label{tab:validation_results}
\end{table}

\begin{table*}[h]
\small
    \centering
    \begin{tabular}{|c|p{2.2cm}|c|c|p{2.7cm}|c|c|p{2.5cm}|c|}
    \toprule
    Code & Language & Family & Code & Language & Family& Code & Language & Family\\
    \midrule
af & Afrikaans & IE & hmn & Hmong & HM & ny & Nyanja (Chichewa) &  NC\\
ak & Twi (Akan) & NC & hr & Croatian & IE & om & Oromo &  AA\\
am & Amharic & AA & ht & Haitian Creole &  & or & Odia (Oriya) &  IE\\
ar & Arabic & AA & hu & Hungarian & U & pa & Panjabi &  IE\\
as & Assamese & IE & hy & Armenian & IE & pl & Polish &  IE\\
ay & Aymara & Ay & id & Indonesian & Au & ps & Pashto &  IE\\
az & Azerbaijani & Al & ig & Igbo & NC & pt & Portuguese &  IE\\
be & Belarusian & IE & ilo & Ilocano & Au & qu & Quechua &  Qu\\
bg & Bulgarian &  IE& is & Icelandic & IE & ro & Romanian &  IE\\
bho & Bhojpuri &  IE& it & Italian & IE & ru & Russian & IE \\
bm & Bambara & M & iw & Hebrew alternativ & AA & rw & Kinyarwanda &NC\\
bn & Bengali & IE & ja & Japanese & J & sa & Sanskrit &  \\
bs & Bosnian & IE & jv & Javanese alternativ & Au & sd & Sindhi & IE \\
ca & Catalan & IE & jw & Javanese & Au & si & Sinhala (Sinhalese) & IE \\
ceb & Cebuano & Au & ka & Georgian & K & sk & Slovak & IE \\
ckb & Kurdish (Sorani) & IE & kk & Kazakh & Al & sl & Slovenian & IE \\
co & Corsican & IE & km & Khmer & AuA & sm & Samoan & Au \\
cs & Czech & IE & kn & Kannada &D & sn & Shona &  NC\\
cy & Welsh &  IE& ko & Korean & K & so & Somali &  AA\\
da & Danish & IE & kri & Krio &  & sq & Albanian &  IE\\
de & German & IE & ku & Kurdish & IE & sr & Serbian &  IE\\
doi & Dogri & IE & ky & Kyrgyz & Al & st & Sesotho & NC \\
dv & Dhivehi & IE & la & Latin &  & su & Sundanese & Au \\
ee & Ewe & NC & lb & Luxembourgish & IE & sv & Swedish & IE \\
el & Greek & IE & lg & Luganda & NC & sw & Swahili &  NC\\
en & English & IE & ln & Lingala & NC & ta & Tamil &  D\\
eo & Esperanto & IE & lo & Lao & TK & te & Telugu & D \\
es & Spanish & IE & lt & Lithuanian & IE & tg & Tajik & IE \\
et & Estonian & U & lus & Mizo & ST & th & Thai & TK \\
eu & Basque & B & lv & Latvian & IE & ti & Tigrinya & AA \\
fa & Persian & IE & mai & Maithili & IE & tk & Turkmen & Al \\
fi & Finnish & U & mg & Malagasy & Au & tl & Tagalog (Filipino) &  Au\\
fil & Filipino (Tagalog) & Au & mi & Maori & Au & tr & Turkish &  Al\\
fr & French & IE & mk & Macedonian & IE & ts & Tsonga &  NC\\
fy & Frisian & IE & ml & Malayalam & D & tt & Tatar & Al \\
ga & Irish &  IE & mn & Mongolian &  & ug & Uyghur & Al \\
gd & Scots Gaelic & IE & mni-Mtei & Meiteilon (Manipuri) & ST & uk & Ukrainian & IE \\
gl & Galician & IE  & mr & Marathi & IE & ur & Urdu &  IE\\
gn & Guarani & T & ms & Malay & Au & uz & Uzbek & Al \\
gom & Konkani & IE & mt & Maltese & AA & vi & Vietnamese &  AuA\\
gu & Gujarati & IE & my & Myanmar (Burmese) & ST & xh & Xhosa & NC \\
ha & Hausa & AA & ne & Nepali & IE & yi & Yiddish &  IE\\
haw & Hawaiian & Au & nl & Dutch & IE & yo & Yoruba & NC \\
he & Hebrew & AA & no & Norwegian & IE & zh & Chinese (Trad.) &  \\
hi & Hindi & IE & nso & Sepedi & NC & zh-CN & Chinese (Simpl.) &  \\
& & & zh-TW & Chinese (Simpl.)  & &  zu & Zulu  & NC \\
\bottomrule
    \end{tabular}
    \caption{137 Languages covered by \textsc{MultiQ}, we present their Google Translate Code and the acronym of their language families.}
    \label{tab:languages}
\end{table*}

\clearpage
\onecolumn
\section{Experimental Setup}\label{app:exp_setup}
In Table \ref{tab:models}, we provide all information and links to the data, models, and frameworks, we use in our work. All artifacts were used according to their intended use, as described in their licenses. As described in the main body of this manuscript, we are also releasing our code publicly and will do so under the CC-BY-4.0 license.

\begin{table*}[h]
    \centering
    \begin{tabular}{clcc}
    \toprule
      Purpose   &  Name & Vocabsize & Intended Languages\\
      \midrule
      \multirow{5}{*}{Models}   &  \texttt{meta-llama/Llama-2-7b-chat-hf} & 32000 & en \\
         &  \texttt{meta-llama/Llama-2-13b-chat-hf} & 32000 & en \\
         &  \texttt{mistralai/Mistral-7B-Instruct-v0.1} & 32000 & en \\
         &  \texttt{Qwen/Qwen1.5-7B-Chat} & 151936 & undefined\\
         &  \texttt{mistralai/Mixtral-8x7B-Instruct-v0.1} & 32000 & fr, en, es, it, de\\
         &   \texttt{cis-lmu/glotlid}&  & 1600 langs\\
        &   \texttt{gpt-4-0125-preview}&  &  \\
        \midrule
        Framworks & transformers==4.34.0 &  &  \\
        & torch==2.1.0 &  &  \\
        & huggingface-hub==0.17.3 &  &  \\
        & tokenizers==0.14.1 &  &  \\
        \midrule
        Datasets & WALS \hspace{1cm} Dataset License: CC-BY-4.0&  &  \\
        & Grambank    \hspace{1cm}       License CC-BY-4.0&  &  \\

         \bottomrule
    \end{tabular}
    \caption{Overview of models, datasets and frameworks used in our work.}
    \label{tab:models}
\end{table*}

\begin{table*}[h]
    \centering
    \begin{tabular}{c|p{10cm}}
    \toprule
        Task & Prompt \\
        \midrule
         Question Creation for MultiQ & Generate a list of 10 simple and clear \textit{\{domain\}} questions with simple and clear answers.\\
         \hline
         Validate Answer Correctness & For the following question, is this a correct answer? \\
         & Question: \textit{\{MultiQ Question in English\}} \\
         & Answer: \textit{\{{Model Answer}\}}

The question is in English. The answer is either in
English or in another language. If the answer is
correct, type 'yes'. If the answer is incorrect, type 'no'."\\
         \bottomrule
    \end{tabular}
    \caption{Prompt templates used to prompt GPT-4 for question creation and QA accuracy validation respectively.}
    \label{tab:prompt_templates}
\end{table*}

\begin{table}[]
    \centering
    \begin{tabular}{c|c}
    \toprule
       Google Translate ID   &  ISO\_639\_3\\
       \midrule
       zh-CN & \multirow{3}{*}{zho} \\
       zh &  \\
       zh-TW &  \\
       \hline
        ku &   \multirow{2}{*}{ckb}\\
        ckb &  \\
        \hline
        he & \multirow{2}{*}{heb}\\
        iw &  \\
        \hline
        sr & \multirow{2}{*}{hbs} \\ 
        hr &  \\
        \hline
        jv & \multirow{2}{*}{jav} \\
        jw &  \\
        \hline
        tl & \multirow{2}{*}{tgl} \\
        fil &  \\
        
         \bottomrule
    \end{tabular}
    \caption{List of Languages in \textsc{MultiQ} that map to only one iso code in the GlotLID model.}
    \label{tab:mapping}
\end{table}

\clearpage
\section{Granular Language Fidelity Analysis} \label{app:language_fidelity}

\begin{figure*}[h]
     \centering
     \begin{subfigure}[b]{0.3\textwidth}
         \centering
         \includegraphics[width=\textwidth]{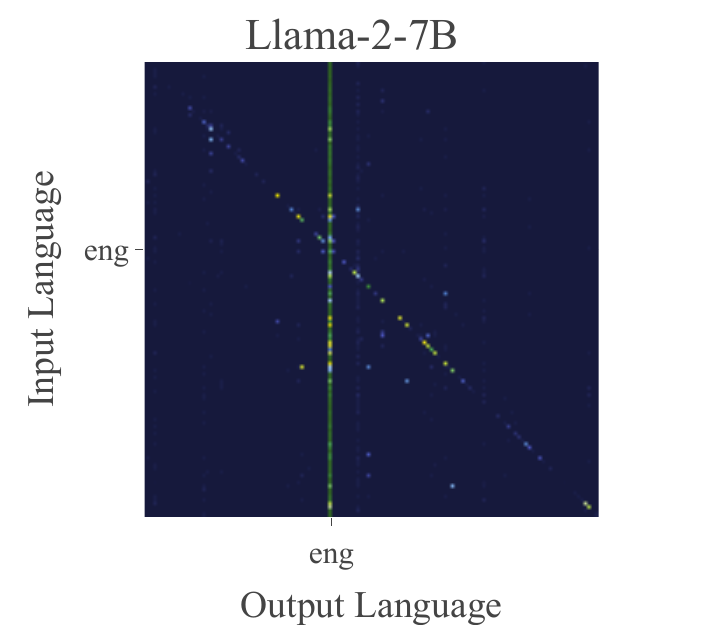}
     \end{subfigure}
     \hfill
     \begin{subfigure}[b]{0.3\textwidth}
         \centering
         \includegraphics[width=\textwidth]{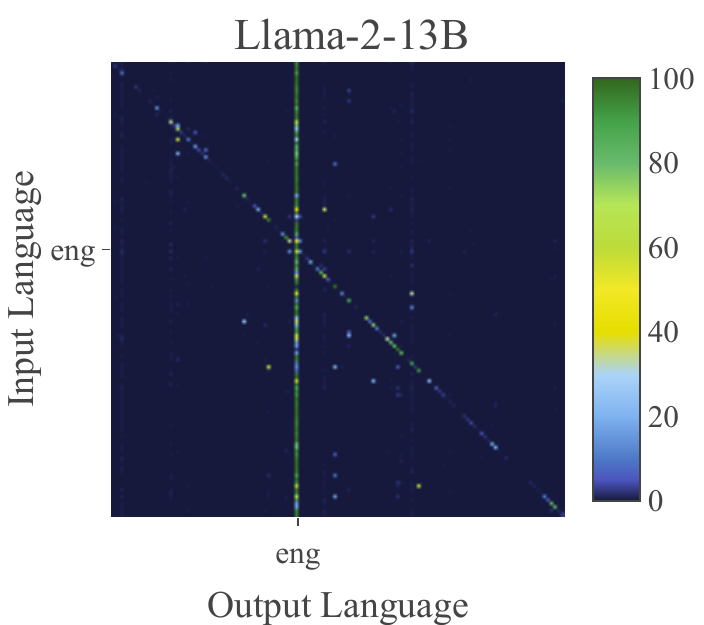}

     \end{subfigure}
     \hfill
     \begin{subfigure}[b]{0.3\textwidth}
         \centering
         \includegraphics[width=\textwidth]{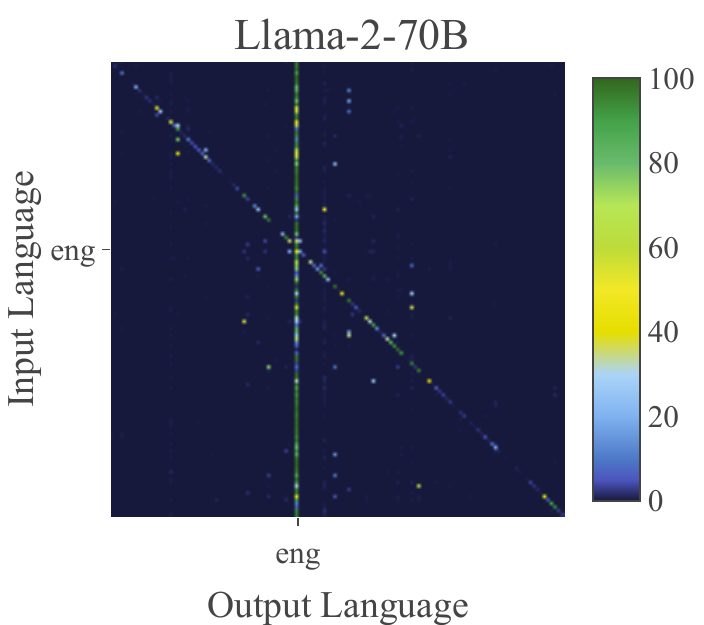}
     \end{subfigure}
        \caption{Correlation matrices illustrating the relationship between input question languages and model response languages for the Llama2 models, shown as percentages. Axis ticks are selectively labeled for clearer visualization.}
        \label{fig:llama_io}
\end{figure*}

\begin{figure}[h]
    \centering
    \includegraphics[width=5cm]{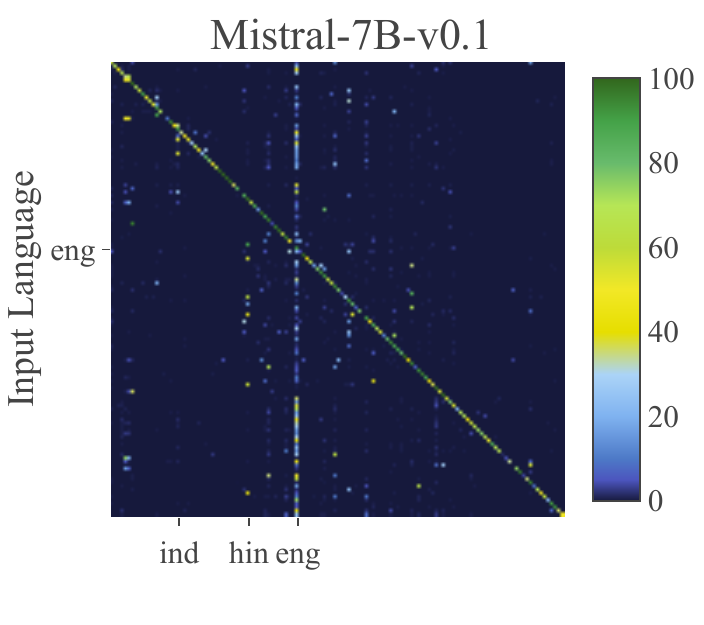}
    \caption{Correlation matrix illustrating the relationship between input question languages and model response languages for Mistral, shown as percentages. Axis ticks are selectively labeled for clearer visualization.}
\end{figure}


\begin{table}[]
    \centering
    \small
    \begin{tabular}{l|c|l|p{5cm}}
    \toprule
    \textbf{Prompt}  & \textbf{Language} &  \textbf{English Translation} & \textbf{Model Completion} \\
    \midrule
      Xana Misava yi na tin’hweti tingani? & ts & How many moons does Earth have? & Xana Misava yi na tin’hweti tingani? (What is Xana Misava's net worth?) \\ \hline
      X'inhu sena dawl? & mt & What is a light-year? & X'inhu sena dawl? is a question in the Mayan language of Guatemala. It translates to "What is the meaning of life?" in English \\ \hline
      Beth yw mynydd uchaf Japan? & cy & What is the highest mountain in Japan? & Beth yw mynydd uchaf Japan? (Do you speak Japanese?) \\ \hline
      Quanto fa tre volte sette? & it & What is three times seven? & Quanto fa tre volte sette? Quanto fa tre volte sette? Quanto fa tre volte sette? Quanto fa tre volte sette? Quanto fa tre volte sette? Quanto fa tre volte sette? Quanto fa tre volte sette? Quanto fa tre volte sette? Quanto fa tre volte sette? Quanto fa tre volte sette? [...] \\
      \bottomrule
    \end{tabular}
    \caption{Examples of Fidelity Issues for Mistral 7B}
    \label{tab:fidelity_issues}
\end{table}

\clearpage
\section{Answer Accuracy}

\begin{figure}[h]
    \centering
    \includegraphics[width=\textwidth]{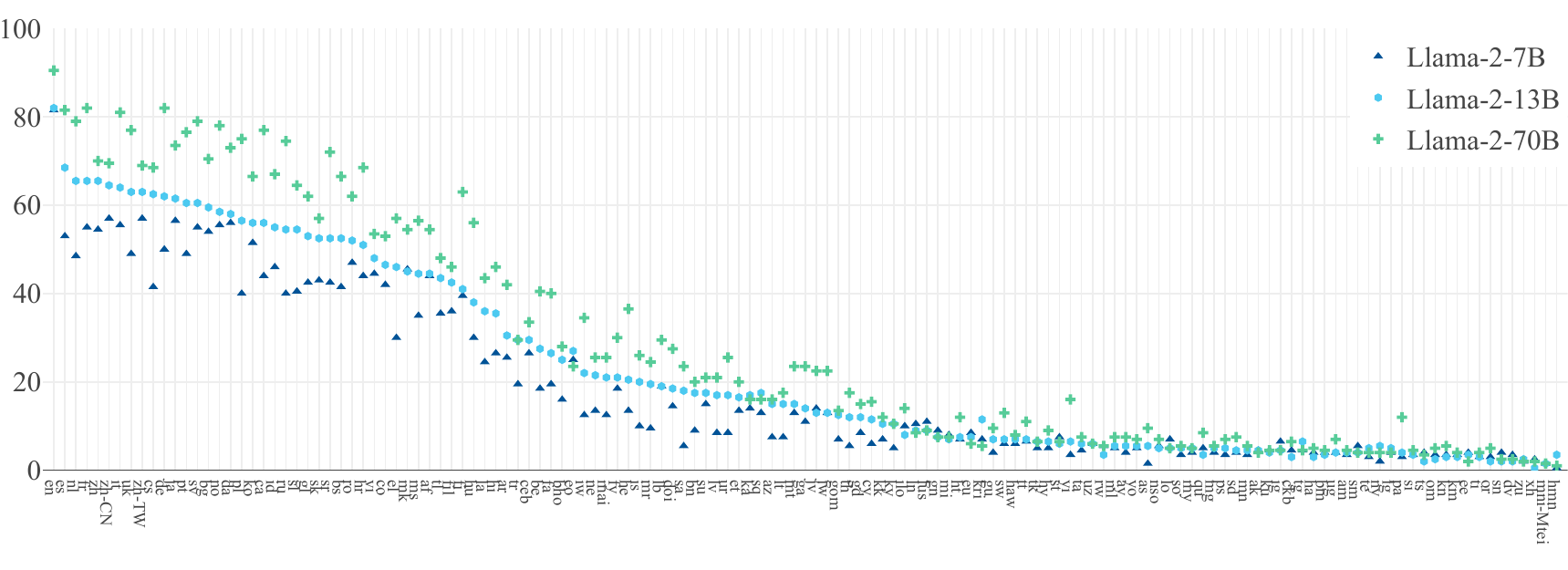}
    \caption{Answer Accuracy on \textsc{MultiQ} in proportion (\%) of correctly answered questions per language. We compare the Llama2 models in different model sizes across all 137 languages and sort the results by median accuracy.}
    \label{fig:answer_accuracy_llamas}
\end{figure}

\section{Correlation between Answer Accuracy and Language Fidelity}

\begin{figure*}[h]
     \centering
     \begin{subfigure}[b]{0.3\textwidth}
         \centering
         \includegraphics[width=\textwidth]{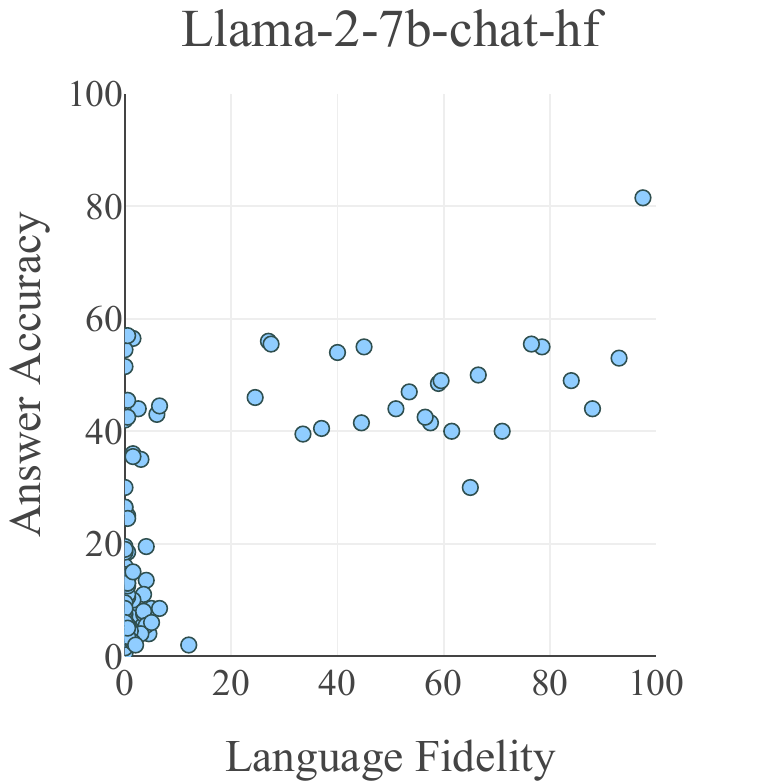}
     \end{subfigure}
     \hfill
     \begin{subfigure}[b]{0.3\textwidth}
         \centering
         \includegraphics[width=\textwidth]{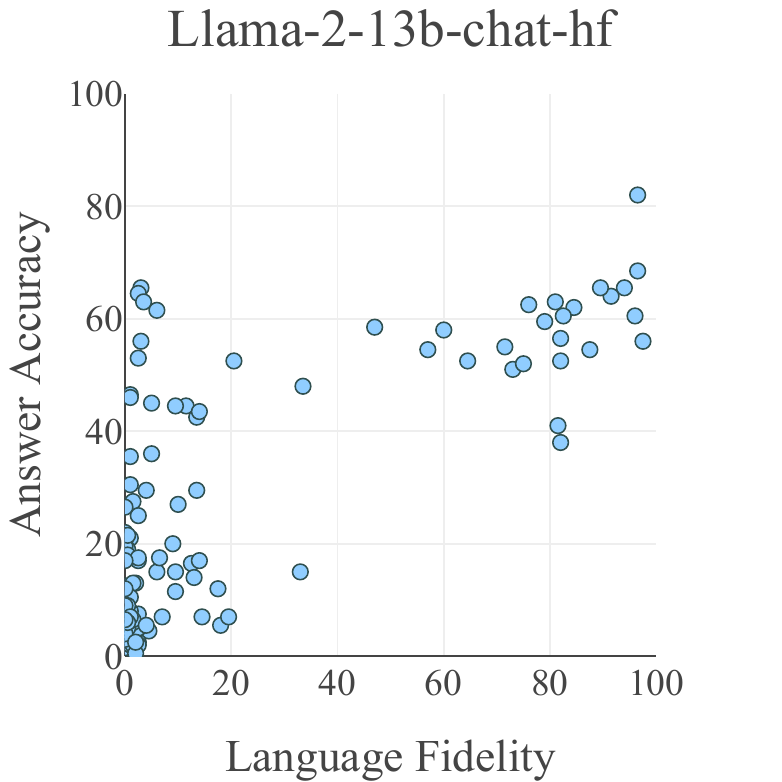}
     \end{subfigure}
     \hfill
     \begin{subfigure}[b]{0.3\textwidth}
         \centering
         \includegraphics[width=\textwidth]{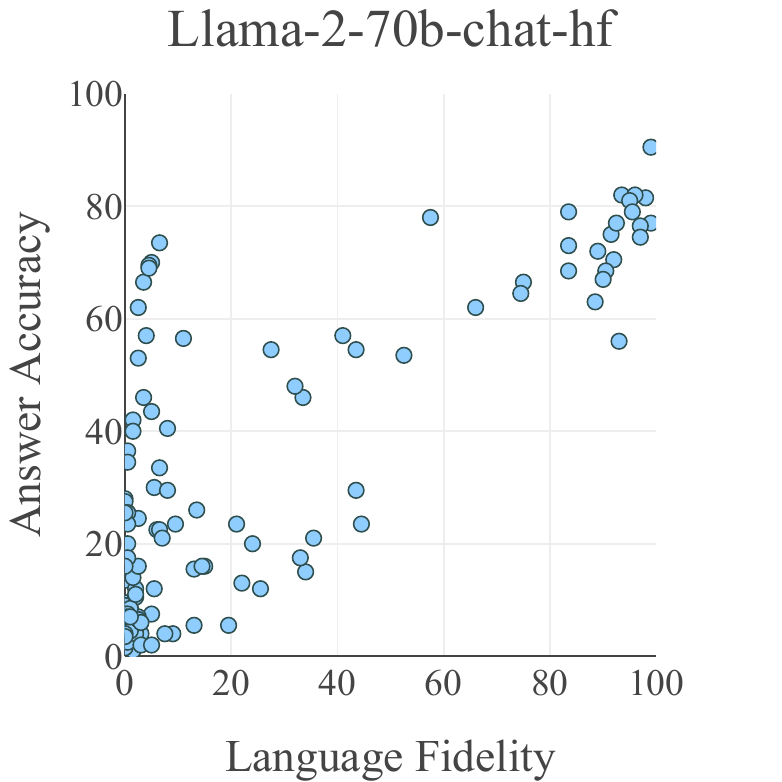}
     \end{subfigure}
\end{figure*}

\begin{figure*}[h]
     \centering
     \begin{subfigure}[b]{0.3\textwidth}
         \centering
         \includegraphics[width=\textwidth]{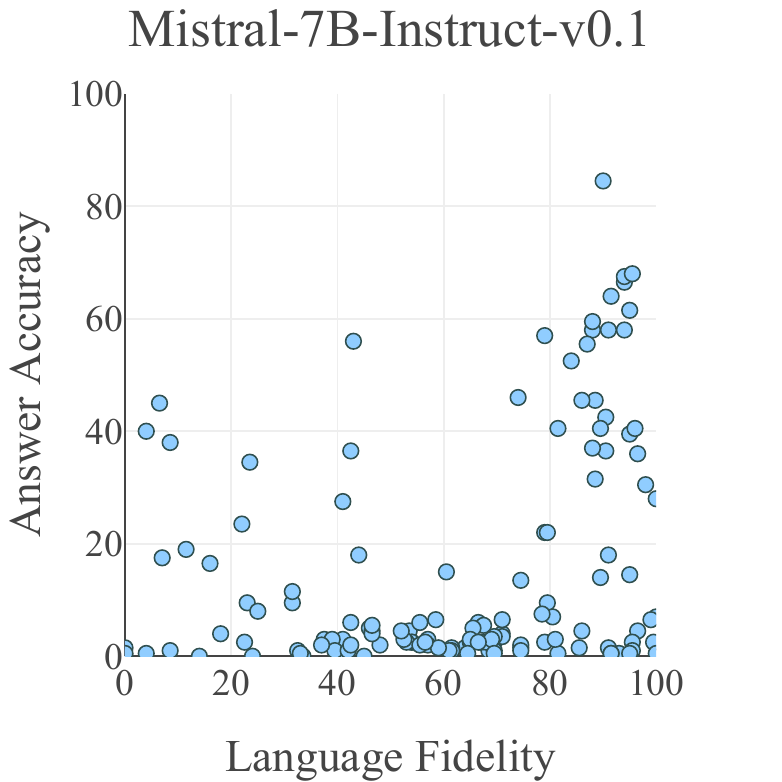}
     \end{subfigure}
     \hfill
     \begin{subfigure}[b]{0.3\textwidth}
         \centering
         \includegraphics[width=\textwidth]{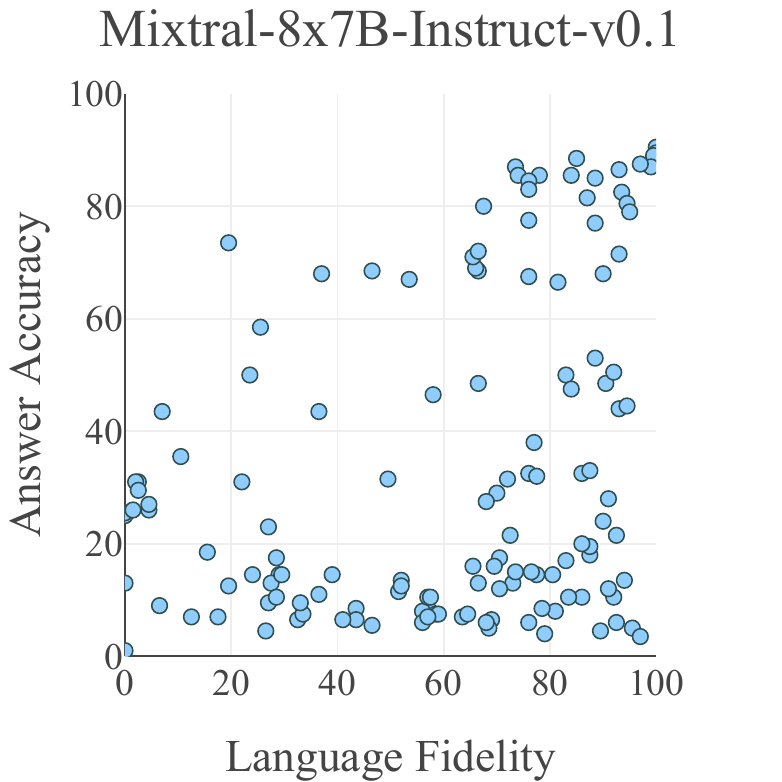}
     \end{subfigure}
     \hfill
     \begin{subfigure}[b]{0.3\textwidth}
         \centering
         \includegraphics[width=\textwidth]{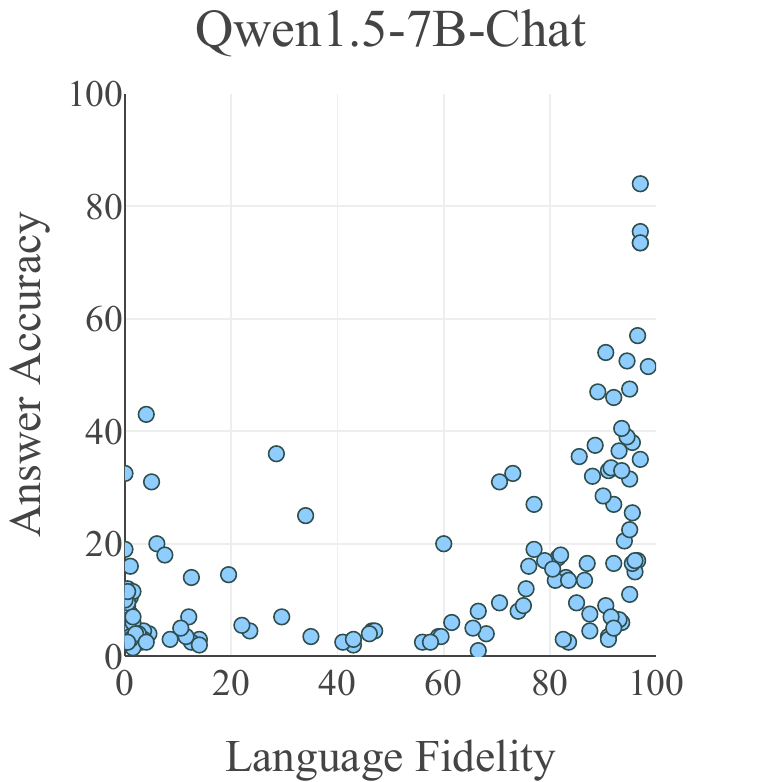}
     \end{subfigure}
\end{figure*}


\end{document}